\documentclass{article}
\usepackage{spconf,amsmath,graphicx}
\usepackage{multirow}
\usepackage{flushend}
\usepackage{url}

\title{Detection of Parasitic Eggs from Microscopy Images and the emergence of a new dataset}
\name{\normalsize Perla Mayo$^{\dagger}$, Nantheera Anantrasirichai$^{\dagger}$, Thanarat H. Chalidabhongse$^{\ddag}$, Duangdao Palasuwan$^{\ddag}$ and Alin Achim$^{\dagger}$ \sthanks{This research was possible thanks to the Newton Fund Institutional Links, British Council 623714323.}}
\address{$^{\dagger}$Visual Information Laboratory, 
	University of Bristol, United Kingdom \\
	$^{\ddag}$Chulalongkorn University, Bangkok, Thailand}
\onecolumn

\begin{document}
%
\maketitle
\begin{abstract}
Automatic detection of parasitic eggs in microscopy images has the potential to increase the efficiency of human experts whilst also providing an objective assessment. The time saved by such a process would both help ensure a prompt treatment to patients, and off-load excessive work from experts' shoulders. Advances in deep learning inspired us to exploit successful architectures for detection, adapting them to tackle a different domain. We propose a framework that exploits two such state-of-the-art models. Specifically, we demonstrate results produced by both a Generative Adversarial Network (GAN) and Faster-RCNN, for image enhancement and object detection respectively, on microscopy images of varying quality. The use of these techniques yields encouraging results, though further improvements are still needed for certain egg types whose detection still proves challenging. As a result, a new dataset has been created and made publicly available, providing an even wider range of classes and variability.
\end{abstract}
\begin{keywords}
Parasitic Egg Detection, GAN, FasterRCNN, Microscopy Images, Pix2Pix
\end{keywords}
\section{Introduction}
\label{sec:intro}
Parasitic eggs have been considered to be predominantly an issue isolated to tropical and subtropical areas, however, due to modern international travel, it is now not uncommon to detect their presence in countries whose environments typically do not support such parasites \cite{garcia2018parasitesgastrointestinal, parasiticeggwhatsgoingon}. The effects produced by these parasites depend not only on the parasitic species in question but can also vary from one host to another  \cite{garcia2018parasitesgastrointestinal}. Diarrhoea and abdominal pain are common symptoms patients may present. However, certain parasites can greatly impact the host, for instance, hookworms ingest human blood, which can lead to anaemia  \cite{garcia2018parasitesgastrointestinal}, whilst the Ascaris parasites have shown to obstruct the biliary and pancreatic ducts. An accurate and speedy detection of these parasites in host patients is vital to ensure that symptoms, especially of more serious conditions, can be diminished and prevented. The most common method of identifying parasitic eggs is that of visually inspecting microscopy images \cite{parasiticeggwhatsgoingon}. From examining a snapshot of stool samples, experts can identify the presence of any specific parasite. However, in practice this has proven to be a tedious and time consuming task, requiring an expert to carefully assess acquired images, one by one, attempting to find and accurately identify any eggs present. The evaluation of each sample, in addition of being highly time consuming, is heavily dependant upon the knowledge and experience of the expert in question for a high success rate. In many areas these requirements of both specialised equipment and expert knowledge greatly constrain the availability of such diagnosis. The automation of this process could offer a solution that would both provide quick results and, importantly, be able to serve a wider population where there is a lack of experts and/or resources. 

In this work, we propose a framework whose goal is the accurate identification of parasitic eggs from microscopy images of stool samples. We intentionally do not focus solely on high quality images, as these are often currently unobtainable in areas where such parasites manifest. Instead, our efforts are directed to the design and implementation of a model that provides a high success rate even with images whose acquisition and appearance are far from ideal.

The rest of the manuscript is organised as follows. Section \ref{sec:related} provides a brief review of similar approaches, not necessarily those addressing the same application, but whose work is still relevant for the task and field. Sections \ref{sec:methods} and \ref{sec:setup} present the method, and experimental setup used to evaluate its performance, which is then reported in section \ref{sec:res}. Lastly, we offer a conclusion and future lines of work in section \ref{sec:conclusions}.

\section{Related Work}
\label{sec:related}
Current techniques make use of deep convolutional neural networks following the success of several architectures that address the object detection task. Waithe et. al. \cite{waithe2020fluorescentcells} assess the performance of four state of the art architectures for the detection of fluorescent cells from microscopy images. On the other hand, novel and domain oriented approaches can also be found, for instance, the recent contribution in \cite{wollmann2021deepconsensus}, the Deep Consensus Network (DCN) architecture, aims to detect cell centroids to facilitate its counting. The object detection networks are often composed of a classification-oriented Convolutional Neural Network (CNN) on top of which a mechanism to address multi-scale targets is placed, often a Feature Pyramid Network (FPN) \cite{lin2017fpn}. The output can be further refined by applying Non-Maximum Suppression (NMS) \cite{hosang2017nms} to discard conflicting overlapping bounding boxes.  One of the key points is that these two works fall under the binary setting as there is no need to differentiate the cells to detect. In \cite{wollmann2021deepconsensus} the effect of detection of objects belonging to different classes was investigated. The result of this was a drop in performance as fine grained differentiation is an inherently more challenging task. 

Similar to what we propose in this manuscript, the contributions in \cite{viet2019parasite} and \cite{suwannaphong2021parasiticmsc} made use of the Faster-RCNN architecture for the detection of the egg parasites, both of them achieving outstanding performance. Of particular interest is the work in \cite{suwannaphong2021parasiticmsc} as the dataset used for this study was acquired using low-cost devices, such that the methodology could be deployed to areas in which the resources, including the experts, might not be available. As a result, the image quality of such dataset is rather poor. Furthermore, in real-life scenarios as the one described here, there is no guarantee that an ideal or optimal image acquisition could be followed, and thus a wider variability unaccounted for could occur. This could be due to the use of a different microscope, changes in lightning that could result in changes of brightness and colour, as well as blurring artefacts due to motion.

 A common way to address image degradation consists in adding a pre-processing step prior to network inference. Enhanced images would then be fed into the object detection model, aiming for an improved performance. The enhancement can also be done by deep architectures, such as Pix2Pix \cite{CHEN2020101743} or CycleGAN \cite{9506694}, for a set of paired or unpaired images, respectively. The work done in \cite{wang2022gansuperresolutionmicroscope} explores this by using both a CycleGAN and a Pix2Pix architecture, the former being used to create paired data for the training of the latter, such that image super resolution can be achieved from an unpaired dataset. 

\section{Detecting parasitic eggs}
\label{sec:methods}

\begin{figure}[htb]
    \begin{minipage}[b]{\linewidth}
      \centering
      \centerline{\includegraphics[width=0.4\linewidth]{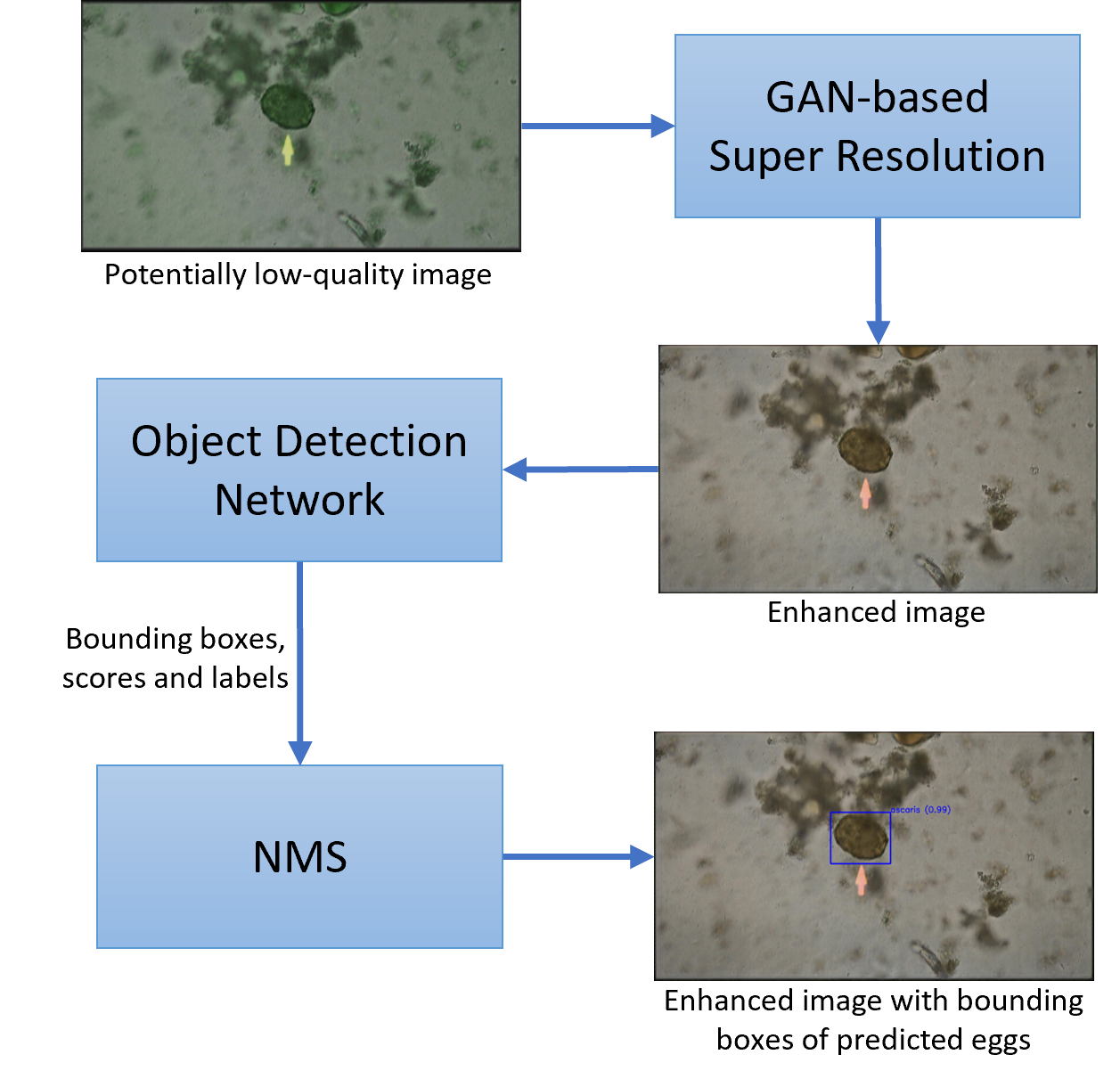}}
    \end{minipage}
    \caption{Block diagram of proposed framework: a two step process in which a given image is initially enhanced by a GAN-based architecture, in this paper, Pix2Pix \cite{isola2017pix2pix}, which is then fed to the object detection network (Faster-RCNN\cite{ren2015fasterrcnn}). The bounding boxes are refined via NMS to then provide the output.}
    \label{fig:block_diagram}
\end{figure}

Because of its proven performance, our proposed framework relies on deep learning architectures in a two-step sequence. An initial step is in charge of enhancing the images prior to being fed to the object detection architecture. This pre-processing is carried out by a Generative Adversarial Network (GAN) architecture that has been trained to learn to transform images from a low- to high-resolution domain. The object detection itself is done by a Faster-RCNN \cite{ren2015fasterrcnn} model using ResNet50 \cite{resnet} as backbone. A block diagram of this approach is shown in Figure \ref{fig:block_diagram}. To choose a suitable architecture for preprocessing, we explore the behaviour of a couple of variations of the GAN model, specifically, we tested both Pix2Pix \cite{isola2017pix2pix} and CycleGAN \cite{CycleGAN2017} independently to enhance the microscopy images available in the dataset prior to their processing for the detection of the egg.

In contrast to the image enhancement done in \cite{wang2022gansuperresolutionmicroscope}, we make use only of either Pix2Pix or CycleGAN, trained by artificially generating the low resolution images. In \cite{wang2022gansuperresolutionmicroscope}, CycleGAN is used to generate training data for Pix2Pix, whereas our approach does not require this as we can generate the paired data the detector network requires to transform low- to high-resolution images. The enhancement module is trained to enhance the input images by denoising, deblurring, as well as removing colour jitter, normalising contrast and brightness. 




\section{Experimental Setup}
\label{sec:setup}

\subsection{Dataset}
To evaluate the framework described in section  \ref{sec:methods}, we apply the method on the parasite egg dataset collected at the Chulalongkorn University from Thailand. Three experts checked each image and the majority vote was used as the label. The data consists of images acquired using a variety of devices and under different environment conditions. For instance, different microscopes as well as lenses have been used. In addition to the images from the microscope, there is an additional collection of images acquired with a phone camera pointing to the microscope eyepiece. The appearance of these images varies among all these different acquisition devices. It was this data collection what highlighted the need of methods that could generalise well to the various settings during image acquisition and that could impact the performance of a given model. Thus, we explored additional techniques that could enable the transformation of the input data in such a way that the architecture would not exhibit a drop in performance.

\begin{figure}[htb]
    \begin{minipage}[b]{0.30\linewidth}
      \centering
      \centerline{\includegraphics[width=5cm]{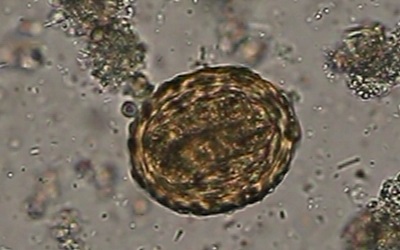}}
      \centerline{(a)  Ascaris }
      \centerline{lumbricoides (AL)}\medskip
    \end{minipage}
    \hfill
    \begin{minipage}[b]{.30\linewidth}
      \centering
      \centerline{\includegraphics[width=5cm]{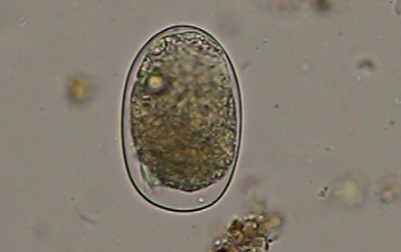}}
      \centerline{(b) Hoowkworm}
      \centerline{(HW)}\medskip
    \end{minipage}
    \hfill
    \begin{minipage}[b]{0.30\linewidth}
      \centering
      \centerline{\includegraphics[width=5cm]{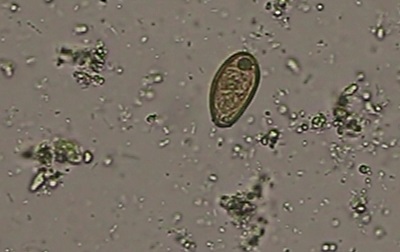}}
      \centerline{(c) Opisthorchis}
      \centerline{viverrine (OV)}\medskip
    \end{minipage}
    \hfill
    \begin{minipage}[b]{0.45\linewidth}
      \centering
      \centerline{\includegraphics[width=5cm]{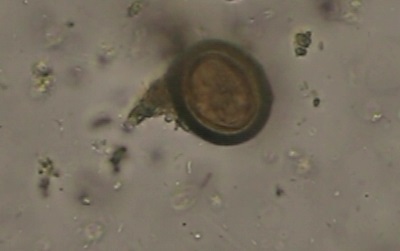}}
      \centerline{(d) Taenia spp. (TS)}
      \centerline{}\medskip
    \end{minipage}
    \hfill
    \begin{minipage}[b]{0.45\linewidth}
      \centering
      \centerline{\includegraphics[width=5cm]{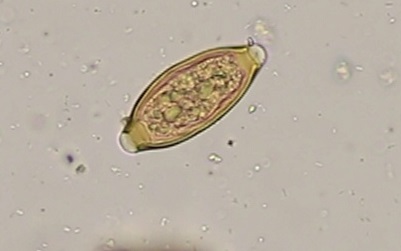}}
      \centerline{(e) Trichuris trichiura}
      \centerline{(Tri)}\medskip
    \end{minipage}
    \caption{Parasitic eggs appearance under a microscope. The features are mostly well defined, however, clutter and motion artifacts can still be observed, as shown for egg (d) Taenia spp.}
    \label{fig:parasiticeggs_examples}
\end{figure}

The parasitic egg dataset contains a total of 2,907 images and 5 classes, comprising 558 from Ascaris lumbricoides (AL), 550 from Hoowkworm (HW), 549 from Opisthorchis viverrine (OV), 551 from Taenia spp. (TS) and 699 from Trichuris trichiura (Tri). Their appearance in microscopy images is shown in Fig. \ref{fig:parasiticeggs_examples}. Some features are clearly defined in these images, whilst some of them present some blurring or changes in lightning conditions. Depending on the microscope used, there could also be a difference in resolution, colour saturation and contrast. The debris shown on the background also largely vary between images.

\subsection{Training procedure}
\subsubsection{Enhancement module}
The enhancement module can be done either with CycleGAN or Pix2Pix. The main difference between these architectures is that the learning of Pix2Pix requires paired data whilst CycleGAN learns the transformation from unpaired images. The networks are trained independently using the original microscope images as target domain whilst the source domain is obtained by artificially altering the original images. Similar to what is done for data augmentation, the low quality domain contains images suffering from blurring, colour jittering, as well as alterations in contrast and brightness. Standard affine transformations are also used to increase the training samples.

Both Pix2Pix and CycleGAN architectures are trained for 500 epochs as this task is more complex. In addition, the learning is done from scratch, i.e., no pre-trained weights. 


\begin{table*}[htb!]
    \centering
    \small
    \begin{tabular}{|c|c|c|c|c|c|c|c|c|c|c|c|c|}
        \hline
        Settings & \multicolumn{6}{|c|}{Precision}&  \multicolumn{6}{|c|}{Recall}  \\ 
          (Training domain/Testing domain)& AL & HW & OV & TS & Tri.& All & AL & HW & OV & TS & Tri. & All \\
         \hline \hline
         Original / Original & 
            0.95 & 0.91 & 0.85 & 0.98 & 0.84 & 0.90 &
            \textbf{1.00} & \textbf{1.00} & 0.97 & 0.99 & \textbf{1.00} & 0.99 \\ \hline
         Original / Grayscale & 
            0.96 & 0.94 & 0.85 & 0.95 & 0.87 & 0.91 & 
            \textbf{1.00} &\textbf{ 1.00} & 0.95 & \textbf{1.00} & \textbf{1.00} & 0.99 \\ \hline
         Original / Low Quality & 
            0.94 & 0.96 & 0.81 & 0.97 & 0.82 & 0.90 & 
            \textbf{1.00} & 0.99 & 0.94 & 0.99 & \textbf{1.00} & 0.99 \\ \hline
        Original / CycleGAN &
            0.82 & 0.87 & 0.64 & 0.95 & 0.92 & 0.84 & 
            0.86 & 0.56 & 0.49 & 0.95 & 0.88 & 0.75 \\ \hline
         Original / Pix2Pix &
            0.94 & 0.96 & 0.81 & 0.97 & 0.82 & 0.90 & 
            \textbf{1.00} & 0.99 & 0.94 & 0.99 & \textbf{1.00} & 0.99 \\ \hline \hline
        
        Pix2Pix / Original &
            0.98 & 0.96 & 0.89 & 0.91 & 0.93 & 0.95 &
            \textbf{1.00} & \textbf{1.00} & \textbf{1.00} & \textbf{1.00} & \textbf{1.00} & \textbf{1.00} \\ \hline
            
        Pix2Pix / Grayscale &
            \textbf{0.99} & 0.97 & 0.83 & \textbf{0.99} & \textbf{0.99} & \textbf{0.97} &
            \textbf{1.00} & \textbf{1.00} & 0.99 & \textbf{1.00} & 0.99 & 0.99 \\ \hline
            
        Pix2Pix / Low Quality &
            0.98 & 0.98 & 0.94 & \textbf{0.99} & 0.94 & \textbf{0.97} &
            \textbf{1.00} & 0.99 & 0.99 & \textbf{1.00} & \textbf{1.00} & 0.99 \\ \hline
            
         Pix2Pix / CycleGAN &
            0.92 & \textbf{0.99} & \textbf{0.99} & 0.98 & 0.97 & \textbf{0.97} &
            0.71 & 0.48 & 0.49 & 0.20 & 0.89 & 0.61 \\ \hline
        Pix2Pix / Pix2Pix &
            0.98 & 0.98 & 0.94 & \textbf{0.99} & 0.94 & \textbf{0.97} &
            \textbf{1.00} & 0.99 & 0.99 & \textbf{1.00} & \textbf{1.00} & 0.99 \\ \hline
    \end{tabular}
    \caption{Precision and recall for the various settings assessed in this study for the classes Ascaris lumbricoides (AL), Hoowkworm (HW), Opisthorchis viverrine (OV), Taenia spp. (TS) and Trichuris trichiura (Tri).}
    \label{tab:method_performance}
\end{table*}

\subsubsection{Detection module}
To detect the parasitic eggs, a pre-trained version of the Faster-RCNN architecture is used for initial training. In contrast to the enhancement module, a total of 50 epochs proved to be enough to successfully address the task, even with data augmentation. The output of the Faster-RCNN is passed through an NMS module to reduce redundancy in the predictions, using a threshold of 0.5 for the Intersection over Union (IoU).

\pagebreak
\section{Results}
\label{sec:res}
The performance of the framework of this paper is judged by the precision and recall for each of the classes in the dataset. We use the standard definitions of true positive, false positive and false negative in the context of object detection. For this, we use a value of 0.5 as a threshold for the IoU for each one of the bounding boxes. Note that we do not assess for bounding boxes for varying scales or IOU thresholds as this would not be done in a real scenario in which the main goal is the identification of the eggs, thus, it suffices to consider a positive case if the IOU is above 0.5 and negative otherwise.

The performance of the proposed approach has been tested on the following scenarios. Firstly, we trained the Faster-RCNN on the original dataset (i.e., no prior image enhancement) and tested on the original image domain. We then assessed the performance of the trained model by using grayscale images on the test set. Note that the network has not been retrained and thus it is expecting colour images. We decided to carry out this experiment as we noticed in some of the executions of the code the performance could be positively affected by this preprocessing on the test dataset. 

The next set of experiments assumed the input data could exhibit alterations the network had not been trained for and thus was not prepared to process. Once again we applied different transformations on the test input data using a different range of functions and parameters to simulate a different range of variability. We tested the original model on this altered dataset. Lastly, for this trained network, we enhance the altered test dataset before passing it to the detector network, once again, without retraining it.

Finally, we carried out the proposed framework by artificially altering the images in the dataset to then be processed by the enhance module. Once again, the alterations the images present are motion blurring, changes in colour, brightness, contrast, saturation as well as image rotation and horizontal and vertical flipping. These images are used to train the Faster-RCNN architecture. Due to the results obtained with the CycleGAN architecture, we performed this experiment only using Pix2Pix for image enhancement.

The above steps are carried out in a 5-fold cross validation set-up. The metrics are averaged across the 5 folds and reported in Table \ref{tab:method_performance}. 
In general, the performance of the Faster-RCNN trained on the images enhanced by Pix2Pix provide the highest values for precision and recall, with the exception of images pre-processed with CycleGAN, as they often provide the lowest values, especially for recall. 
We noticed that for enhanced images used on the standard Faster-RCNN, the rate of false positives for these two classes increased, suggesting that the image enhancement step might introduce inaccurate information that fools the Faster-RCNN network. This seems to be eliminated if we follow the proposed framework (last row in the table). Moreover, this framework presents improvement in most of the metrics shown in Table \ref{tab:method_performance}, which further encourages the use of this method.

\begin{figure}[htb!]
    \begin{minipage}[b]{0.48\linewidth}
      \centering
      \centerline{\includegraphics[width=6cm]{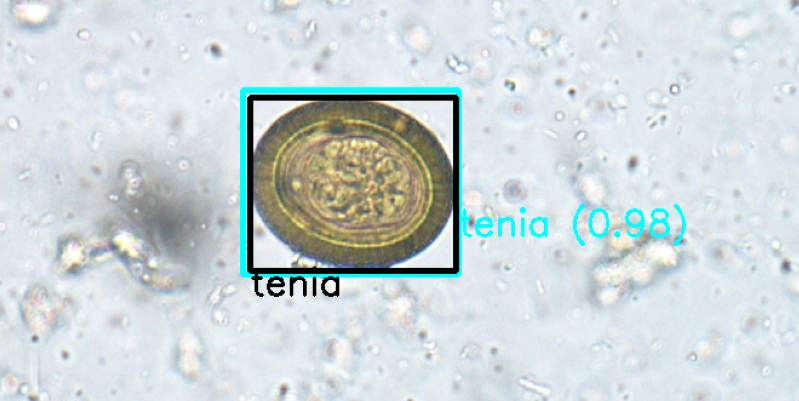}}
      \centerline{(a)  Original}\medskip
    \end{minipage}
    \hfill
    \begin{minipage}[b]{.48\linewidth}
      \centering
      \centerline{\includegraphics[width=6cm]{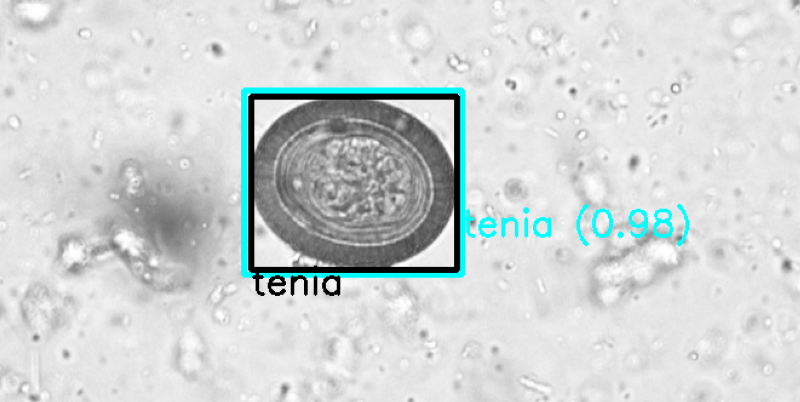}}
      \centerline{(b) Grayscale}\medskip
    \end{minipage}
    \hfill
    \begin{minipage}[b]{0.48\linewidth}
      \centering
      \centerline{\includegraphics[width=6cm]{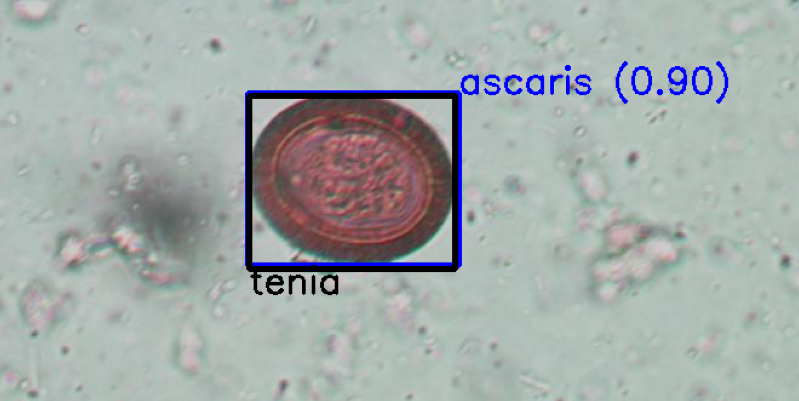}}
      \centerline{(c) Degraded}\medskip
    \end{minipage}
    \hfill
    \begin{minipage}[b]{0.48\linewidth}
      \centering
      \centerline{\includegraphics[width=6cm]{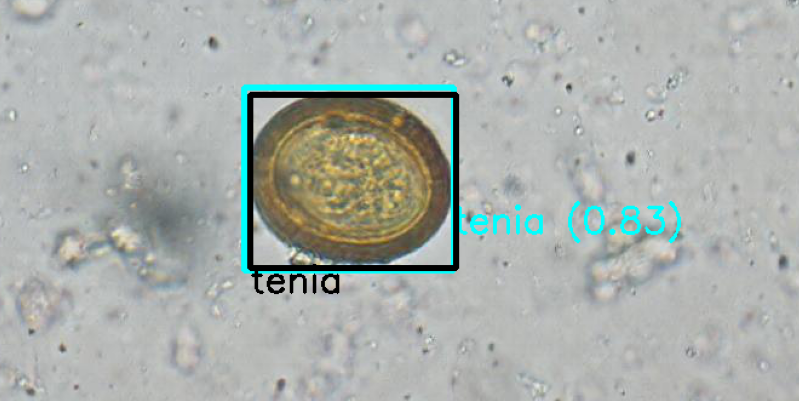}}
      \centerline{(d) Processed with Pix2Pix}\medskip
    \end{minipage}
    \hfill
    \caption{Example of egg detection (Taenia spp.) under different scenarios. a) original image, b) grayscale, b) degraded with colour jittering, and d) after processed with Pix2Pix.}
\label{fig:detection_res}
\end{figure}

We present in Fig. \ref{fig:detection_res} an image sample in the four different versions fed to the egg detector. In some instances, when the image has suffered a considerable level of degradation, the egg detector would fail in the prediction, which can be resolved by the improved image processed by Pix2Pix. 


\section{Conclusions and Future Work}
\label{sec:conclusions}
The proposed approach addresses the detection of parasitic eggs from microscopy images by exploiting state-of-the-art machine learning models. We highlight the importance of a framework that generalises well to different environment conditions as an optimal setting for the image acquisition is not guaranteed. To validate the proposed approach, we inspected the behaviour of what could be considered as a standard deep learning implementation by training the Faster-RCNN network on the original domain and testing it with previously unseen data. The additional incorporation of GANs, aims to improve the sensitivity and specificity of the main framework. We showed that an improvement in performance can be achieved by training the detector on images from a unique domain, which corresponds to the high-quality data generated by the GAN module. 

Future work will explore the use of different object detection architectures under this same framework, aiming to improve speed and performance, this based on findings derived from the assessment of this same task for a different dataset domain \cite{waithe2020fluorescentcells}. 
This would maintain the current structure of the framework, in which image enhancement is achieved independently from the object detection. It has been argued, however, that such enhancement does not necessarily imply an improvement in the classification or detection task \cite{itgandobetter}. Thus, a potential and promising line of work is the investigation of a shared feature domain between low- and high-quality images that could enable an improved performance despite the lack of an image enhancement step for the human reader. By doing so, the framework would be reduced to a unique step in which the transformation of the image would have a direct impact on the learning of the object detection. To encourage further research, we have made publicly available the dataset used for this work in \url{https://bit.ly/3ImvkmC}, as well as helped in organising the Parasitic Egg Detection challenge, which contains an even richer and more challenging dataset. 




\bibliographystyle{IEEEbib}
\bibliography{refs}

\end{document}